\RequirePackage{fix-cm}
\documentclass[twocolumn]{svjour3}          %
\smartqed  %
\usepackage{times}
\usepackage{epsfig}
\usepackage{graphicx}
\usepackage{amsmath}
\usepackage{amssymb}
\usepackage{adjustbox}
\usepackage{mathtools}
\usepackage{xspace}
\usepackage[pagebackref=true,breaklinks=true,letterpaper=true,colorlinks,bookmarks=false]{hyperref}

\newcommand{\eg}{\textit{e.g.}}
\newcommand{\ie}{\textit{i.e.}}

\def\etal{{\em et al.\/}\,}

\newcommand{\prgan}{\textsc{PrGAN}\xspace}
\newcommand{\prgans}{\textsc{PrGANs}\xspace}
\newcommand{\norm}[1]{\left\lVert#1\right\rVert}

\begin{document}

\title{Inferring 3D Shapes from Image Collections using Adversarial Networks%
}

\author{Matheus Gadelha \and
        Aartika Rai  \and
        Subhransu Maji  \and
        Rui Wang
}

\institute{Matheus Gadelha, Aartika Rai, Subhransu Maji, Rui Wang \at
  College of Information and Computer Sciences, University
  of Massachusetts Amherst,   140 Governors Dr, Amherst, MA 01003, USA\\
  \email{\{mgadelha, aartikarai, smaji, ruiwang\}@cs.umass.edu}           %
}

\date{Received: June 10th, 2019}

\maketitle

\begin{abstract}
We investigate the problem of learning a probabilistic distribution over
three-dimensional shapes given two-dimensional views of multiple
objects taken from unknown viewpoints.
Our approach called \emph{projective generative adversarial network}
(\prgan) trains a deep generative model of 3D shapes whose projections
(or renderings) match the distributions of the provided 2D
distribution.
The addition of a \emph{differentiable projection
module} allows us to infer the underlying 3D shape distribution without
access to any explicit 3D or viewpoint annotation during the learning
phase. 
We show that our approach produces 3D shapes of comparable
quality to GANs trained directly on 3D data. %
Experiments also show that the
disentangled representation of 2D shapes into geometry and viewpoint
leads to a good generative model of 2D shapes.
The key advantage of our model is that it estimates 3D shape, viewpoint, and generates novel
views from an input image in a completely unsupervised manner.
We further investigate how the generative models can be improved
if additional information such as depth, viewpoint or part
segmentations is available at training time.
To this end, we present new differentiable projection operators that can be
used by \prgan to learn better 3D generative models.
Our experiments show that our method can successfully leverage extra visual cues
to create more diverse and accurate shapes.
\end{abstract}

\section{Introduction}\label{s:intro}

\begin{figure}[t]
\centering
\includegraphics[width=0.9\linewidth]{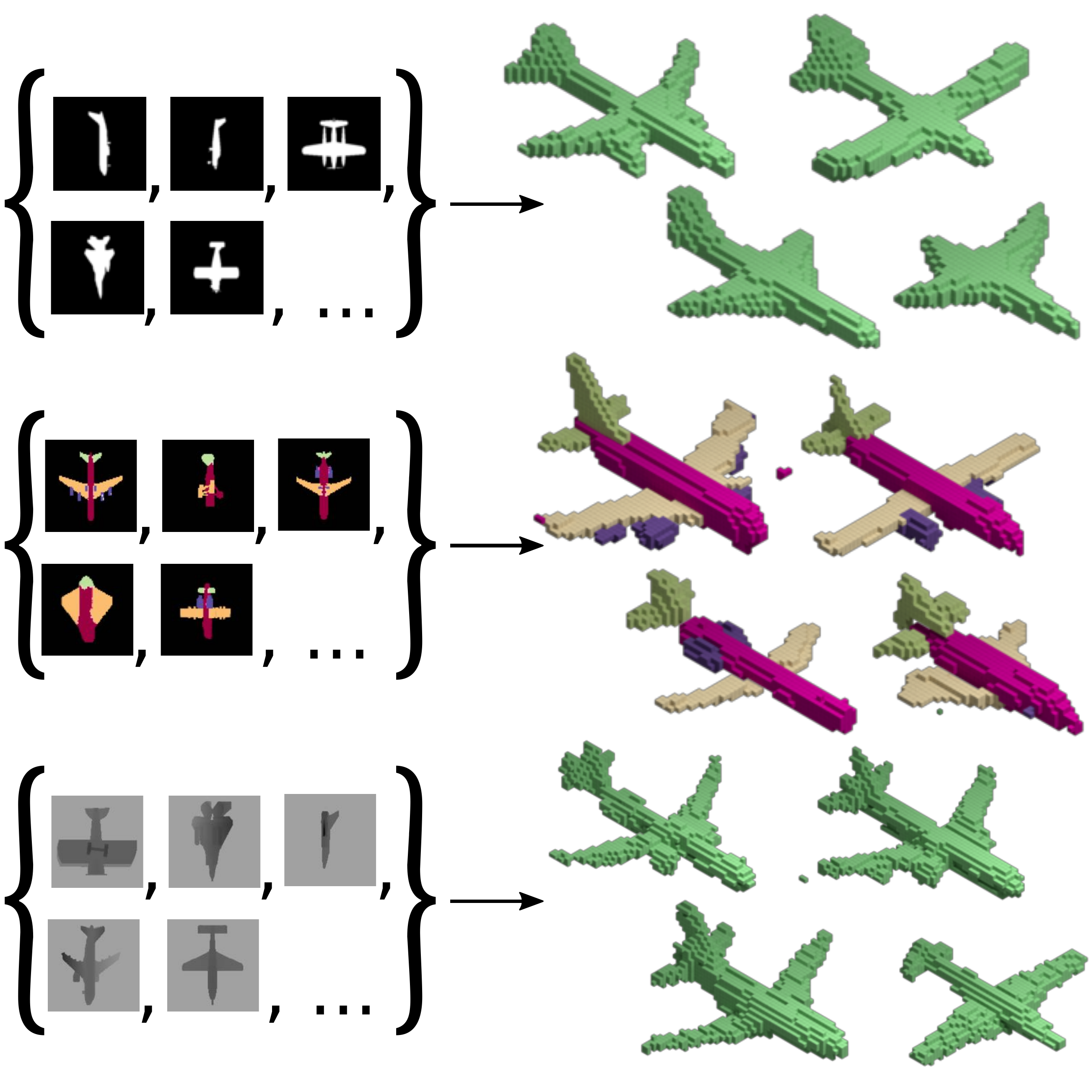}
\caption{\label{f:problem} Our algorithm infers a generative model of
  the underlying 3D shapes given a collection of unlabelled images rendered as
  silhouettes, semantic segmentations or depth maps. 
	To the left, images representing the input dataset.
	To the right, shapes generated by the generative model trained with those images.}
\vspace{-8pt}
\end{figure}

The ability to infer 3D shapes of objects from their 2D views is one
of the central challenges in computer vision.
For example, when presented with a catalogue of different
airplane silhouettes as shown in the top of Figure~\ref{f:problem}, one can mentally infer their
3D shapes by simultaneously reasoning about the shared variability of 
their underlying shape and viewpoint variation. 
In this work, we investigate the problem of learning a generative model of 3D shapes
given a collection of images of an unknown set of objects within a category taken from
an unknown set of views. 
The problem is challenging since one is not provided with the information about
which object instance was used to generate each image, the viewpoint from
which each image was taken, the parametrization of the underlying shape
distribution, or even the number of underlying
instances.
This makes it difficult to apply traditional techniques based on structure
from motion~\cite{hartley2003multiple,blanz1999morphable} or visual
hulls~\cite{laurentini1994visual}.

\begin{figure*}
\centering
\includegraphics[width=\linewidth]{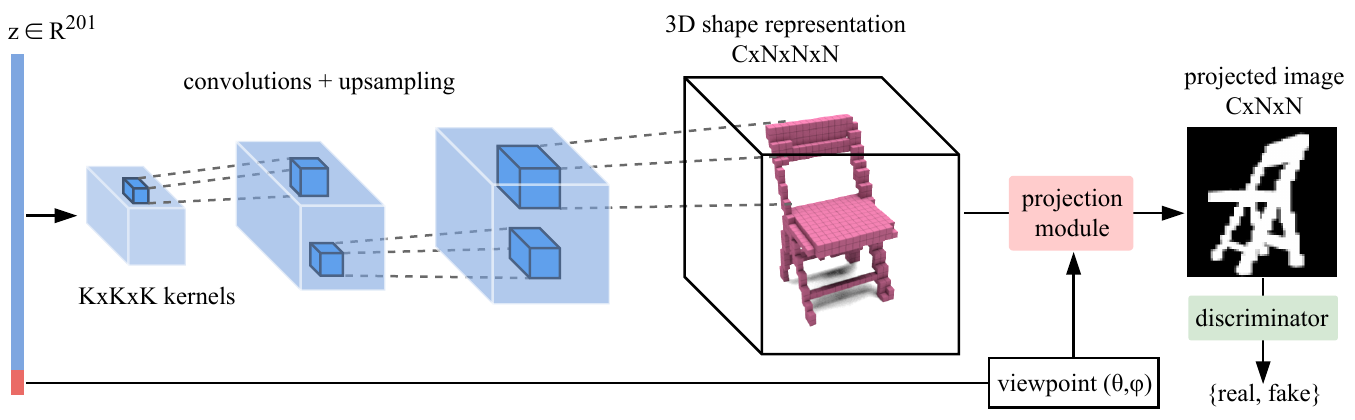}
\caption{\label{fig:prgan-arch} \emph{The PrGAN architecture for generating
  2D silhouettes of shapes factorized into a 3D shape generator and
  viewpoint generator and projection module.} A 3D voxel representation ($\text{C}
  \times \text{N}^3$) and
  viewpoint are independently generated from the input $z$ (201-d
  vector). The projection module renders the voxel shape from a given
  viewpoint $(\theta,\phi)$ to create an image. The discriminator
  consists of 2D convolutional and pooling layers and aims to classify
  if the generated image is ``real'' or ``fake''. The number of
  channels C in the generated shape is equal to one for an
  occupancy-based representation and is equal to the number of parts
  for a part-based representation.}

\end{figure*}

We use the framework of generative adversarial
networks (GANs)~\cite{goodfellow2014generative} and augment the
generator with a \emph{projection module}, as illustrated in Figure~\ref{fig:prgan-arch}.
The generator learns to produce 3D shapes, the projection module renders each shape from
different views, and the adversarial network discriminates real
images from generated ones.
The projection module is a \emph{differentiable renderer} that allows us to map 3D
shapes to 2D images, as well as back-propagate the gradients of 2D
images to 3D shapes.
Once trained, the model can be used to infer 3D shape distributions
from a collection of images (Figure~\ref{f:problem} shows some samples
drawn from the generator), and to infer depth or viewpoint from a single
image, without using any 3D or viewpoint information during learning.
We call our approach \emph{projective generative adversarial network}
~(\prgan).

While there are several cues for inferring the 3D shape from a
single image, we begin with shapes that are rendered as silhouettes.
The motivation is that silhouettes can be easily
extracted when objects are photographed against clear backgrounds, such
as in catalogue images, but nevertheless they contain rich shape
information.
Real-world
images can also be used by removing background and converting them to
binary images. 
Our generative 3D model represents shapes using a voxel representation that indicates
the occupancy of a volume in fixed-resolution 3D grid. 
Our projection module is a feed-forward operator that
renders the volume as an image.
The feed-forward operator is differentiable, allowing
the ability to adjust the 3D volume based on projections. 
Finally, we assume that the distribution over viewpoints
is known (assumed to be uniform in our experiments, but could be any
distribution). 

We then extend our analysis first presented in our earlier
work~\cite{prgan} to incorporate other forms of
supervision and improve the generative models by incoporating advances
in training GANs. 
Additional supervision includes availability of viewpoint information for
each image, depth maps instead of silhouettes, or semantic
segmentations such as part labels during learning.
Such supervision is easier to collect than acquiring full 3D scans of
objects.
For example, one can use a generic object viewpoint
estimator~\cite{su2015render} as weak supervision for our problem.
Similarly semantic parts can be labelled on images directly and
already exist for many object categories such as airplanes, birds,
faces, and people.
We show that such information can be used to improve 3D reconstruction
by using an appropriate projection module.

To summarize our main contributions are as follows: (i) we propose
\prgan, a framework to learn probabilistic distributions over 3D
shapes from a collection of 2D views of objects. We demonstrate its
effectiveness on learning complex shape categories
such as chairs, airplanes, and cars sampled from online shape
repositories~\cite{chang2015shapenet,wu20153d}. 
The results are reasonable, even when views from multiple categories
are combined; (ii) \prgan generates 3D shapes of comparable quality to GANs trained
directly on 3D data~\cite{wu2016learning};
(iii) The learned 3D representation can be used for unsupervised
estimation of 3D shape and viewpoint given a novel 2D shape, 
and for interpolation between two different shapes, (iv) Incoporating
additional cues as weak supervision improves the 3D shapes
reconstructions in our framework.

\section{Related work}\label{s:related}

\paragraph{Estimating 3D shape from image collections.} 
The difficulty of estimating 3D shape can vary widely based on how the
images are generated and the assumptions one can make about the underlying
shapes.
Visual-hull techniques~\cite{laurentini1994visual} can be used to
infer the shape of a particular object given its views from known
viewpoints by computing the intersection of the projected silhouettes. 
When the viewpoint is fixed and the lighting is known, photometric
stereo~\cite{woodham1980photometric} can provide accurate geometry
estimates for rigid and diffuse surfaces.
Structure from motion (SfM)~\cite{hartley2003multiple} can be used to
estimate the shape of \emph{rigid objects} from their views taken from
unknown viewpoints by jointly reasoning about point correspondences
and camera projections. 
Non-rigid SfM can be used to recover shapes from image collections by
assuming that the 3D shapes can be represented using a compact parametric model.
An early example is by Blanz and Vetter~\cite{blanz1999morphable} for
estimating 3D shapes of faces from image collections where each shape
is represented as a linear combination of bases (Eigen shapes). 
However, 3D shapes need to be aligned in a
consistent manner to estimate the bases which can be challenging.
Recently, non-rigid SfM has been applied to categories such as cars
and airplanes by manually annotating a fixed set of keypoints across
instances to provide correspondences~\cite{kar2015category}.
Our work augments non-rigid SfM using a learned 3D shape generator,
which allows us to generalize the technique to categories with diverse
structures \emph{without} requiring correspondence annotations.
Our work is also related to recent work of Kulkarni
\etal~\cite{kulkarni2015deep} for estimating a disentangled
representation of images into shape, viewpoint, and lighting variables
(dubbed ``inverse graphics networks"). However, the shape
representation is not explicit, and the approach requires the ability
to generate training images while varying one factor at a time.

\paragraph{Inferring 3D shape from a single image.} Optimization-based
approaches put priors on geometry, material, and light and estimate
all of them by minimizing the reconstruction error when
rendered~\cite{land1971lightness,barrow1978recovering,BarronTPAMI2015}.
Recognition-based methods have been used to estimate geometry of outdoor scenes~\cite{hoiem2005geometric,saxena2005learning}, indoor environments~\cite{eigen2015predicting,schwing2012efficient}, and objects~\cite{andriluka2010monocular,savarese20073d}. 
More recently, convolutional networks have been trained to generate views of 3D objects given their attributes and camera parameters~\cite{dosovitskiy2015learning}, to generate 3D shape given a 2D view of the object~\cite{tatarchenko2016multi}, and to generate novel views of an object~\cite{zhou2016view}. Most of these approaches are trained in a fully-supervised manner and require 3D data or multiple views of the same object during training.

\paragraph{Generative models for images and shapes.} Our work builds
on the success of GANs for generating images across a wide range of
domains~\cite{goodfellow2014generative}. 
Recently, Wu \etal \cite{wu2016learning} learned a generative model of
3D shapes using GAN equipped with 3D convolutions.
However, the model was trained with aligned 3D shape data.
Our work aims to solve a more difficult question of learning a 3D-GAN
from 2D images. 
Several recent works are in this direction. 
Rezende \etal~\cite{rezende2016unsupervised} show results for 3D
shape completion for simple shapes when views are provided, 
but require the viewpoints to be known and the generative models are
trained on 3D data. 
Yan \etal~\cite{yan2016perspective} learn a mapping from an image
to 3D using multiple 
projections of the 3D shape from known viewpoints and object
identification, i.e., which images correspond to the same object. 
Their approach employs a 3D volumetric decoder and optimizes a loss that measures the 
overlap of the projected volume on the multiple silhouettes from known viewpoints, 
similar to a visual-hull reconstruction. 
Tulsiani \etal~\cite{drcTulsiani17} learn a model to map images to 3D
shape provided with color images or silhouettes of objects taken from
known viewpoints using a ``ray consistency'' approach similar to our
projection module.
Our method on the other hand does not assume known viewpoints or
object associations of the silhouettes making the problem considerably
harder.
More similar to our setup, Henderson and Ferrari~\cite{henderson18} propose a method to learn
a generative model of 3D shapes from a set of images without viewpoint supervision.
However, their approach uses a more constrained shape representation -- sets of blocks or deformations
in a subdivided cube -- and other visual cues, like lighting configuration and normals.

\paragraph{Differentiable renderers.} 
Our generative models rely on a
differentiable projection module to incorporate image-based supervision.
Since our images are rendered as silhouettes, the process
can be approximated using differentiable functions composed of spatial
transformations and projections as described in Section~\ref{s:method}. 
However, more sophisticated differentiable renders, such as~\cite{nmr,Liu18,dmc}, that take into
account shading and material properties could provide richer
supervision or enable learning from real images.
However, these renderers rely on mesh-based or surface-based representations which is 
challenging to generate due to their unstructured nature.
Recent work on generative models of 3D shapes with point
cloud~\cite{lin18,mrt,gadelha3d17,psg,atlasnet} or multiview~\cite{lun17,tatarchenko2016multi} representations
provide a possible alternative to our voxel based approach that we aim
to investigate in the future.

\section{Method}\label{s:method}
\begin{figure*}
  \centering
  \includegraphics[width=0.85\linewidth]{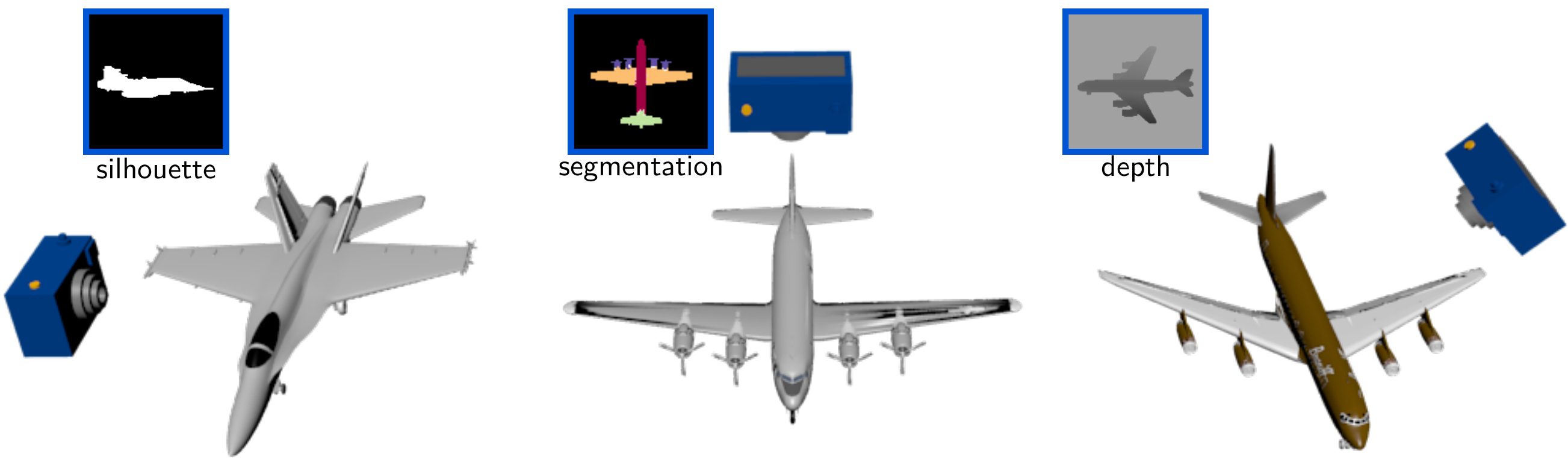}
  \caption{\label{fig:projection} 
    The input to our model consists of multiple renderings of
    different objects taken from different viewpoints.
    Those image are \emph{not} annotated with identification or viewpoint information.
    Our model is able to handle images from objects rendered as
    silhouettes (left), semantic segmentation maps (middle) or
    depth maps (right).
}
\vspace{-8pt}
\end{figure*}

Our method builds upon the GANs proposed in
Goodfellow~\etal~\cite{goodfellow2014generative}.
The goal of a GAN is to train a generative model in an
adversarial setup.
The model consists of two parts: a \emph{generator} and a
\emph{discriminator}.
The generator $G$ aims to transform samples drawn from a simple
distribution ${\cal P}$ that appear to have been sampled from the
original dataset.
The discriminator $D$ aims to distinguish samples generated
by the generator from real samples (drawn from a data distribution ${\cal D}$).
Both the generator and the discriminator are trained jointly by optimizing:

\begin{equation}\label{eqn:gan}
\min_{G}\max_{D} \mathbb{E}_{x\sim{\cal D}} [ \log \left(D(x)\right) ] + \mathbb{E}_{z\sim{\cal P}} [ \log \left(1-D(G(z))\right)].
\end{equation}

Our main task is to train a generative model for 3D shapes without
relying on 3D data itself, instead relying on 2D images from those shapes, without
any view or shape annotation\footnote{We later relax this to
  incorporate additional supervision.}.
In other words, the data distribution consists of 2D images taken from
different views and are of different objects.
To address this issue, we factorize the 2D image generator into a 3D
shape generator (${\cal G}_{3D}$), viewpoint generator $(\theta,\phi)$, and a
projection module ${\cal P_{\theta,\phi}}$ as seen in Figure~\ref{fig:prgan-arch}.
The challenge is to identify a representation suitable for a diverse set of shapes
and a differentiable projection module to create final 2D images and
enable end-to-end training. 
We describe the architecture employed for each of these next. 

\paragraph{3D shape generator ($G_{3D}$).}
The input to the entire generator is $z \in \mathbb{R}^{201}$ with
each dimension drawn independently from a uniform distribution
$\text{U}(-1,1)$.
Our 3D shape generator ${\cal G}_{3D}$ transforms the first 200 dimensions of $z$ to a
$N\times N \times N$ voxel representation of the shape.
Each voxel contains a value $v\in[0, 1]$ that represents its occupancy.
The architecture of the 3D shape generator is inspired by the
DCGAN~\cite{radford2015unsupervised} and 3D-GAN~\cite{wu2016learning}
architectures. 
It consists of a several layers of 3D convolutions, upsampling, and
non-linearities shown in Figure~\ref{fig:prgan-arch}.
The first layer transforms the 200 dimensional vector to a $256\times
4 \times 4 \times 4$ vector using a fully-connected layer.
Subsequent layers have batch normalization and ReLU layers between them
and use 3D kernels of size $5\times5\times5$.
At every layer, the spatial dimensionality is increased by a factor of 2 and
the number of channels is decreased by the same factor, except for the last layer whose
output only has one channel (voxel grid).
The last layer is succeeded by a sigmoid activation instead of a ReLU in order
to keep the occupancy values in $[0,1]$.

\paragraph{Viewpoint generator $(\theta,\phi)$.}
The viewpoint generator takes the last dimension of $z \in
\text{U}(-1,1)$ and transforms it to a viewpoint vector
$(\theta,\phi)$.
The training images are assumed to have been generated from 3D models
that are upright oriented along the y-axis, and are centered at the
origin.
Most models in online repositories and the real world satisfy this
assumption (\eg, chairs are on horizontal planes). We generate images
by sampling views uniformly at random from one of eight pre-selected
directions evenly spaced around the y-axis (\ie, $\theta=0$ and
$\phi=0^\circ$, $45^\circ$, $90^\circ$, $...$, $315^\circ$), as seen
in Figure~\ref{fig:projection}.
Thus the viewpoint generator picks one of these directions uniformly
at random.

\paragraph{Projection module ($Pr$).}
The projection module $Pr$ renders the 3D shape from the given
viewpoint to produce an image. 
For example, a silhouette can be
rendered in the following steps.
The first step is to rotate the voxel grid to the corresponding viewpoint.
Let $V : \mathbb{Z}^3 \rightarrow [0,1] \in \mathbb{R}$ be the voxel grid, a function that given
given an integer 3D coordinate $c=(i,j,k)$ returns the occupancy of the voxel centered at $c$.
The rotated version of the voxel grid $V(c)$ is defined as 
$V_{\mathbb{\theta, \phi}} = V(\lfloor{R(c, \mathbf{\theta,\phi})}\rfloor)$,
where $R(c, \mathbf{\theta},\phi)$ is the coordinate obtained by rotating $c$ around the origin
according to the spherical angles $\mathbf{(\theta,\phi)}$. 

The second step is to perform the projection to create an image from the rotated voxel
grid.
This is done by applying the projection operator 
$Pr((i,j),V) = 1 - e^{-\sum_{k}V(i,j,k)}$. 
Intuitively, the operator sums up the voxel occupancy values along
each line of sight (assuming orthographic projection), and applies
exponential falloff to create a smooth and differentiable
function. 
When there is no voxel along the line of sight, the value is
0; as the number of voxels increases, the value approaches 1. Combined
with the rotated version of the voxel grid, we define our final
projection module as:
$Pr_{\theta,\phi}((i,j),V) = 1 - e^{-\sum_{k}V_{\theta,\phi}(i,j,k)}$. 
As seen in Figure~\ref{fig:projection} the projection module can well
approximate the rendering of a 3D shape as a binary silhouette image,
and is differentiable.
In Section~\ref{s:discussion} we present projection modules that
render the shape as a depth image or labelled with part segmentations
using similar projection operations as seen in
Figure~\ref{fig:projection}. 
Thus, the 2D image generator $G_{2D}$ can be written
compositionally as ${G}_{2D} = {Pr}_{(\theta, \phi)} \circ G_{3D}$.

\paragraph{Discriminator ($D_{2D}$).}
The discriminator consists of a sequence of 2D convolutional layers with
batch normalization layer and LeakyReLU activation~\cite{maas2013rectifier} between them. 
Inspired by recent work~\cite{radford2015unsupervised,wu2016learning}, we employ
multiple convolutional layers with stride 2 while increasing the
number of channels by 2, except for the first layer,
whose input has 1 channel (image) and output has 256.
Similar to the generator, the last layer of the discriminator is
followed by a sigmoid activation instead of a LeakyReLU.

\paragraph{Training details.}
We train the entire architecture by optimizing the objective in Equation~\ref{eqn:gan}.
Usually, the training updates to minimize each one of the losses is applied once at each
iteration.
However, in our model, the generator and the discriminator have a considerably different
number of parameters.
The generator is trying to create 3D shapes, while the discriminator is trying to
classify 2D images.
To mitigate this issue, we employ an adaptive training strategy.
At each iteration of the training, if the discriminator accuracy is higher than 75\%,
we skip its training.
We also set different different learning rates for the discriminator and the generator:
$10^{-5}$ and $0.0025$, respectively.
Similarly to the DCGAN architecture \cite{radford2015unsupervised}, we use ADAM with $\beta=0.5$ for the optimization.

\section{Experiments}\label{s:experiments}
In this section we describe the set of experiments to evaluate our method
and to demonstrate the extension of its capabilities.
First, we compare our model with a traditional GAN for the task
of image generation and a GAN for 3D shapes.
We present quantitative and qualitative results.
Second, we demonstrate that our method is able to induce 3D shapes from
unlabelled images even when the collection contains only a single view per object.
Third, we present 3D shapes induced by our model from a variety of
categories such as airplanes, cars, chairs, motorbikes, and
vases. Using the same architecture, we show how our model is able to
induce coherent 3D shapes when the training data contains images mixed
from multiple categories.
Finally, we show applications of our method in predicting 3D shape
from a novel 2D shape, and performing shape interpolation.

\begin{figure*}[t]
\setlength{\tabcolsep}{0pt}
\centering
\begin{tabular}{cccccccc}
\includegraphics[width=.12\linewidth]{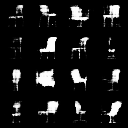} &
\includegraphics[width=.12\linewidth]{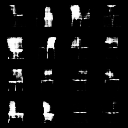} &
\includegraphics[width=.12\linewidth]{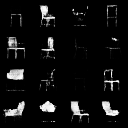} &
\includegraphics[width=.12\linewidth]{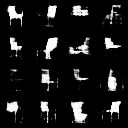} \hfill&
\includegraphics[width=.12\linewidth]{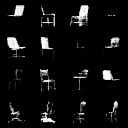} &
\includegraphics[width=.12\linewidth]{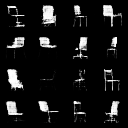} &
\includegraphics[width=.12\linewidth]{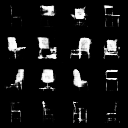} &
\includegraphics[width=.12\linewidth]{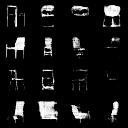} \\
	\multicolumn{4}{c}{(a) Results from 2D-GAN.} \vspace{4pt} &
	\multicolumn{4}{c}{(a) Results from \prgan.} \vspace{4pt}
\end{tabular}
\caption{\label{fig:validation2} Comparison between 2D-GAN~\cite{goodfellow2014generative} and our \prgan model for image generation on the chairs dataset. Refer to Figure~\ref{fig:generate-shapes} third row, left column for samples of the input data.}
\end{figure*}

\paragraph{Input data.} We generate training images synthetically using 3D shapes available in the
ModelNet~\cite{wu20153d} and ShapeNet~\cite{chang2015shapenet} databases. Each category contains a few hundred to thousand shapes. We render each
shape from 8 evenly spaced viewing angles with orthographic projection to produce binary images. Hence our assumption is that the viewpoints of the training images (which are unknown to the network) are uniformly distributed. If we have prior knowledge about the viewpoint distribution (e.g. there may be more frontal views than side views), we can adjust the projection module to incorporate this knowledge. To reduce aliasing, we
render each image at $64\times64$ resolution and downsample to $32\times32$. We have found that this generally improves
the results. Using synthetic data allows us to easily perform controlled experiments to analyze our method. 
It is also possible to use real images downloaded from a search engine as discussed in Section \ref{s:discussion}.

\subsection{Results}
We quantitatively evaluate our model by comparing its ability to generate
2D and 3D shapes.
To do so, we use 2D image GAN similar to DCGAN~\cite{radford2015unsupervised} and a 3D-GAN
similar to the one presented at \cite{wu2016learning}.
At the time of this writing the implementation of \cite{wu2016learning} is not public yet, therefore we
implemented our own version.
We will refer to them as 2D-GAN and 3D-GAN, respectively.
The 2D-GAN has the same discriminator architecture as the \prgan, but the generator contains a sequence of 2D transposed convolutions instead of 3D ones, and
the projection module is removed.
The 3D-GAN has a discriminator with 3D convolutions instead of 3D ones.
The 3D-GAN generator is the same of the \prgan, but without the projection module.

The models used in this experiment are chairs from ModelNet dataset~\cite{wu20153d}.
From those models, we create two sets of training data: voxel grids and images.
The voxel grids are generated by densely sampling the surface and inside of each mesh, and binning the sample points into $32\times32\times32$ 
grid.
A value 1 is assigned to any voxel that contains at least one sample point, and 0 otherwise.
Notice that the voxel grids are only used to train the 3D-GAN, while the images are used to train the 2D-GAN and our \prgan.

\begin{figure*}[t]
\setlength{\tabcolsep}{0pt}
\centering
\begin{tabular}{cccc|cccc}
\includegraphics[width=.12\linewidth]{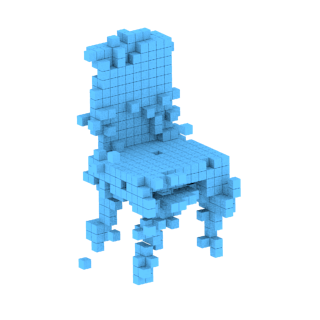} &
\includegraphics[width=.12\linewidth]{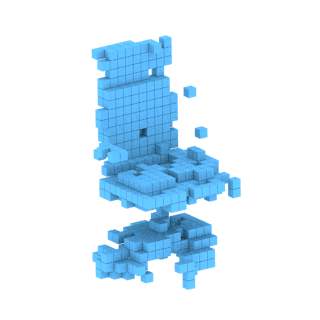} &
\includegraphics[width=.12\linewidth]{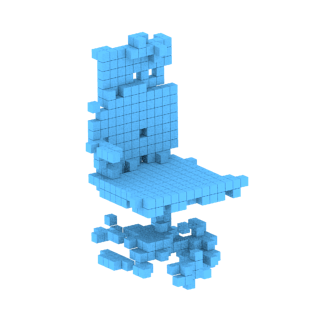} &
\includegraphics[width=.12\linewidth]{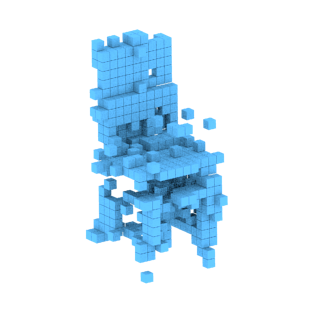} &
\includegraphics[width=.12\linewidth]{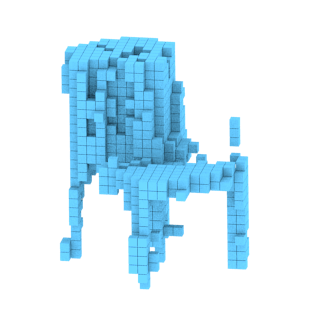} &
\includegraphics[width=.12\linewidth]{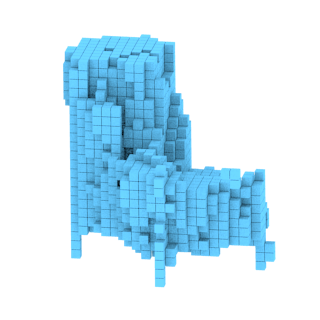} &
\includegraphics[width=.12\linewidth]{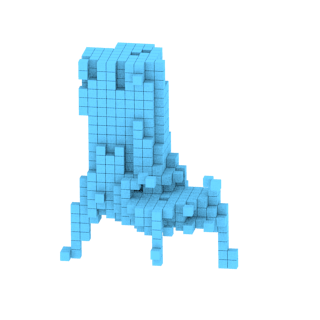} &
\includegraphics[width=.12\linewidth]{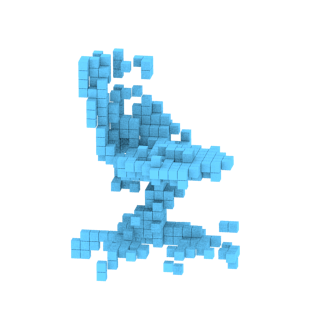} \\
\includegraphics[width=.12\linewidth]{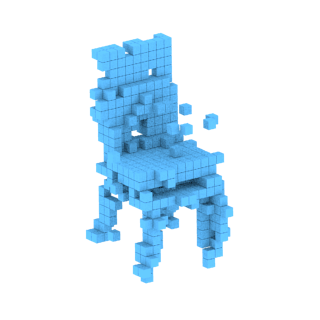} &
\includegraphics[width=.12\linewidth]{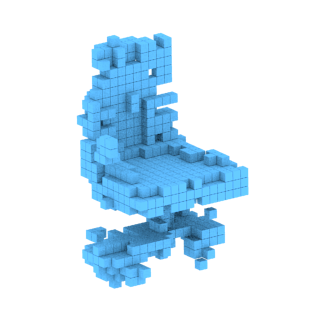} &
\includegraphics[width=.12\linewidth]{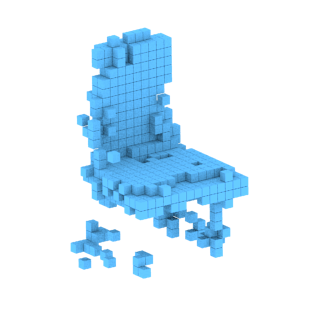} &
\includegraphics[width=.12\linewidth]{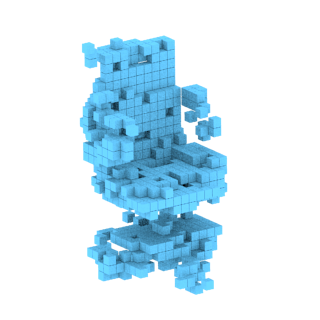} &
\includegraphics[width=.12\linewidth]{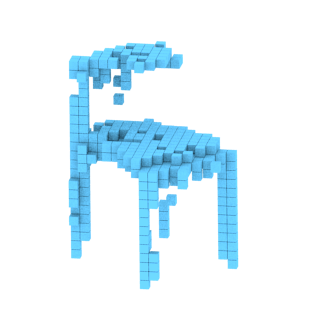} &
\includegraphics[width=.12\linewidth]{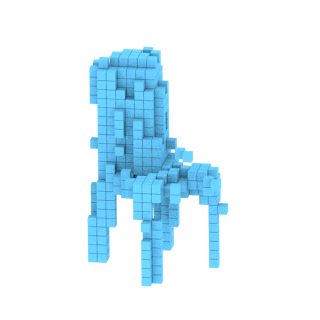} &
\includegraphics[width=.12\linewidth]{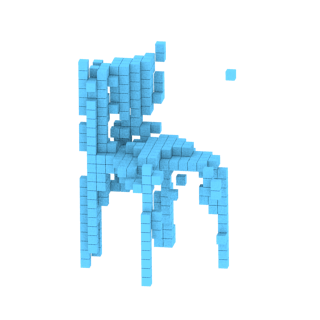} &
\includegraphics[width=.12\linewidth]{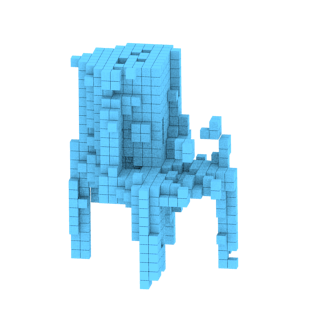} \\
\multicolumn{4}{c}{(a) Results from 3D-GAN.} &
\multicolumn{4}{c}{(a) Results from \prgan.}\\
\end{tabular}
\caption{\label{fig:validation3} Comparison between 3D-GAN~\cite{wu2016learning} and our \prgan for 3D shape generation. The 3D-GAN is trained on 3D voxel representation of the chair models, and the \prgan is trained on images of the chair models (refer to Figure~\ref{fig:generate-shapes} third row).}
\end{figure*}

Our quantitative evaluation is done by taking the Maximum Mean Discrepancy (MMD)~\cite{gretton2006kernel} 
between the data created by the generative models and the training data.
We use a kernel bandwidth of $10^{-3}$ for images and $10^{-2}$ for voxel grids.
The training data consists of 989 voxel grids and 7912 images.
To compute the MMD, we draw 128 random data points from each one of the generative models.
The distance metric between the data points is the hamming distance divided by the dimensionality of the data.
Because the data represents continuous occupancy values, we binaritize them by using a threshold of $0.001$ for images or voxels created
by \prgan, and $0.1$ for voxels created by the 3D-GAN.

\begin{figure}[t]
\includegraphics[width=0.95\linewidth]{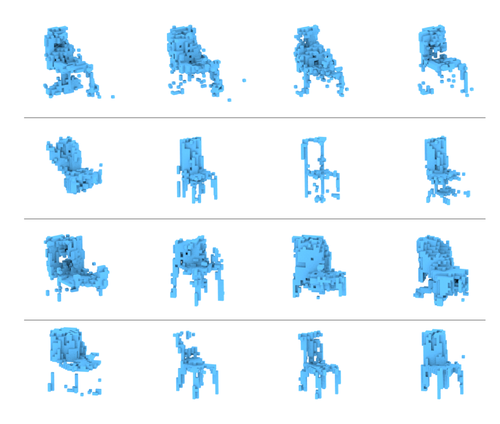}
\caption{\label{fig:varying_views} Shapes generated from \prgan by varying the number of views per object in the training data. From the top row to the bottom row, the number of views per object in the training set are 1, 2, 4, and 8 respectively.}
\vspace{-8pt}
\end{figure}

Results show that for 2D-GAN, the MMD between the generated images and the training data is \textbf{90.13}. For \prgan, the MMD is \textbf{88.31}, which is slightly better quantitatively than 2D-GAN. Figure~\ref{fig:validation2} shows a qualitative comparison. The results are visually very similar. For 3D-GAN, the MMD between the generated voxel grids and the training voxel grids is \textbf{347.55}. For \prgan, the MMD is \textbf{442.98}, which is worse compared to 3D-GAN. This is not surprising as 3D-GAN is trained on 3D data, while \prgan is trained on the image views only. Figure~\ref{fig:validation3} presents a qualitative comparison. In general \prgan has trouble generating interior structures because the training images are binary, carries no shading information, and are taken from a limited set of viewing angles. Nonetheless, it learns to generate exterior structures reasonably well.

\subsubsection{Varying the number of views per model} 
In the default setting, our training data is generated by samples 8 views per
object. 
Note that we do not provide the association between views and instances
to the generator.
Here we study the ability of our method in the more challenging case
where the training data contains fewer number of views per object. 
To do so, we generate a new training set that contains only 1 randomly
chosen view per object and use it to train \prgan. We then repeat the
experiments for 2 randomly chosen views per object, and also 4. The
results are shown in Figure~\ref{fig:varying_views}. The 3D shapes that
\prgan learns becomes increasingly better as the number of views
increases. Even in the single view per object case, our method is able
to induce reasonable shapes.

\begin{figure}[t]
\setlength{\tabcolsep}{0pt}
\begin{tabular}{cccccc}
\includegraphics[width=.166\linewidth]{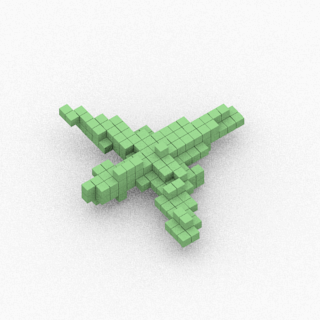} &
\includegraphics[width=.166\linewidth]{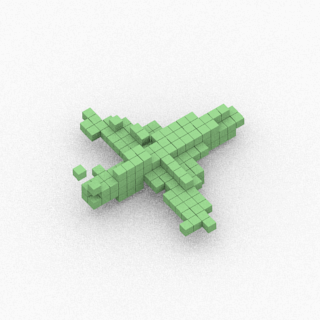} &
\includegraphics[width=.166\linewidth]{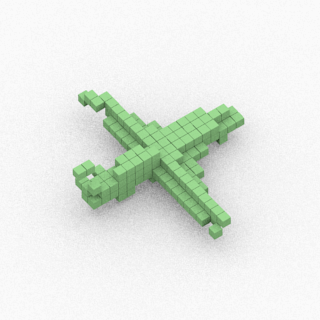} &
\includegraphics[width=.166\linewidth]{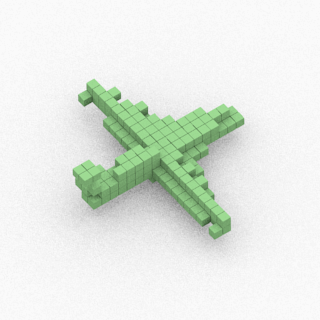} &
\includegraphics[width=.166\linewidth]{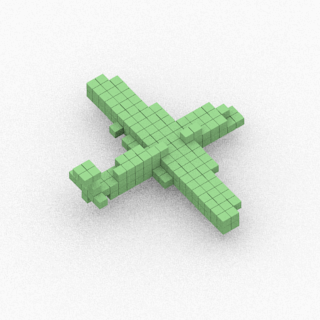} &
\includegraphics[width=.166\linewidth]{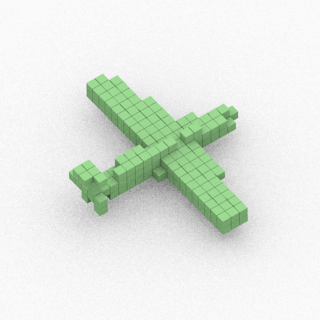} \\
\includegraphics[width=.166\linewidth]{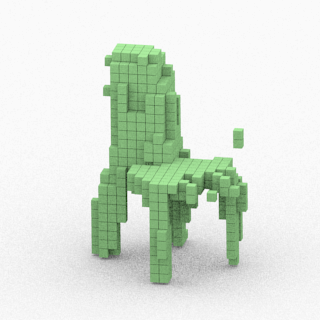} &
\includegraphics[width=.166\linewidth]{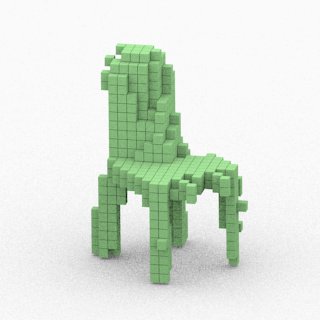} &
\includegraphics[width=.166\linewidth]{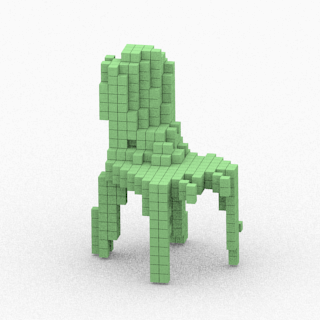} &
\includegraphics[width=.166\linewidth]{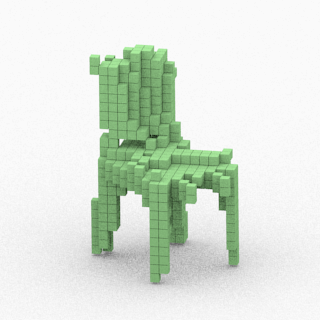} &
\includegraphics[width=.166\linewidth]{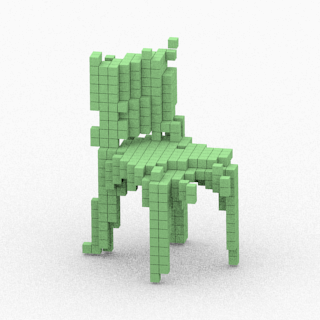} &
\includegraphics[width=.166\linewidth]{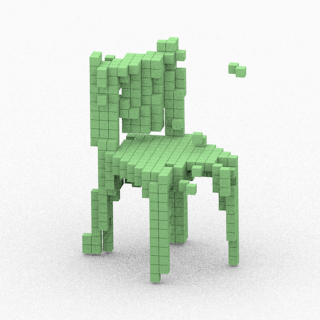} \\
\end{tabular}
\caption{\label{fig:interp} Shape interpolation by linearly interpolating the encodings of the starting shape and ending shape.}
\end{figure}

\subsubsection{Shape interpolation}
Once the generator is trained, any encoding $z$ supposedly generates a
plausible 3D shape, hence $z$ represents a 3D shape manifold. Similar
to previous work, we can interpolate between 3D shapes by linearly
interpolating their $z$ codes. Figure~\ref{fig:interp} shows the
interpolation results for two airplane models and two chair models.

\begin{figure}[t]
	\centering
	\includegraphics[width=0.85\linewidth]{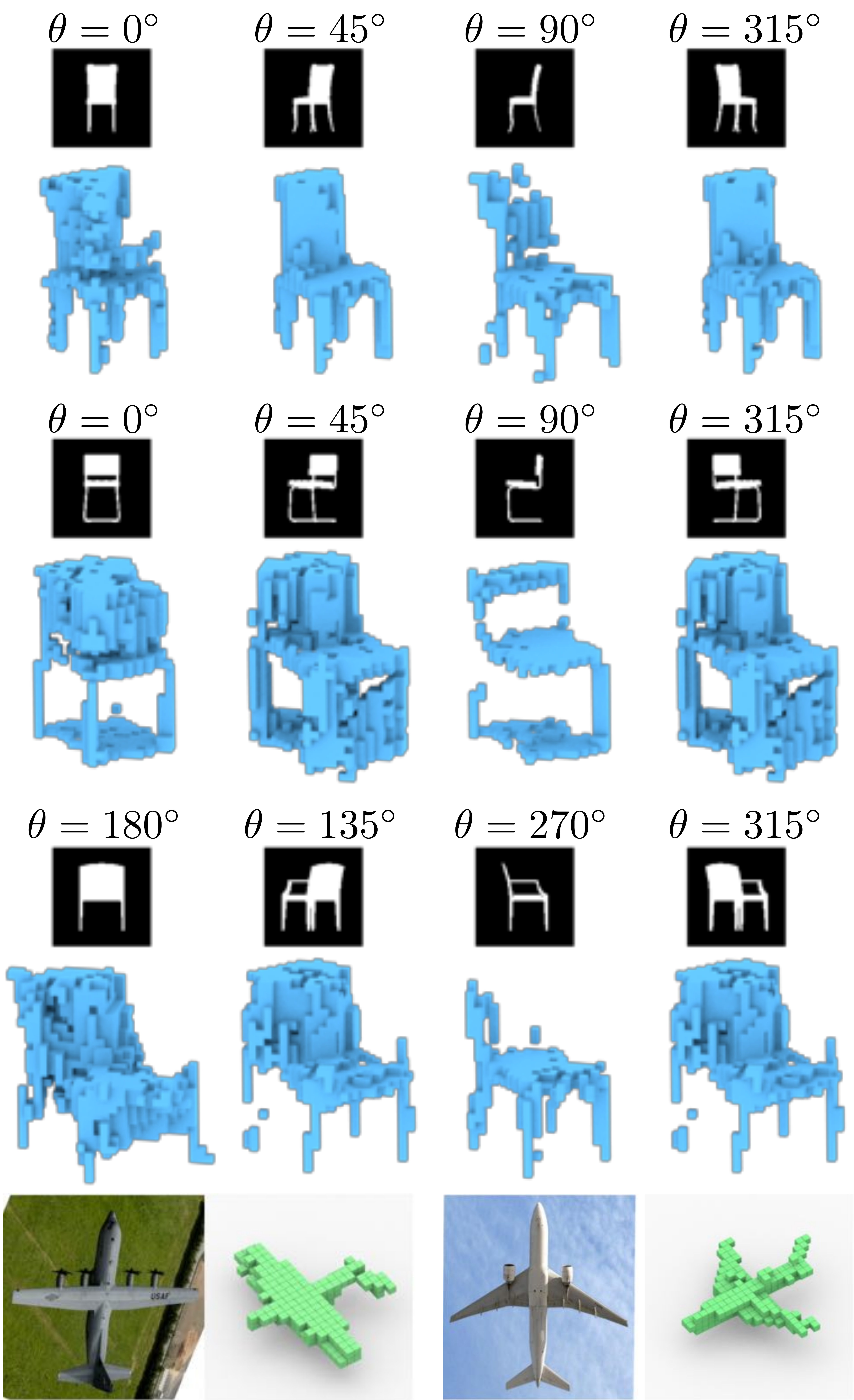}
	\caption{\label{fig:sp} 
	At top 3 rows, the four images are different views of the same chair, with predicted viewpoint on the top. 
	Shapes are different but plausible given the single view.
	In the bottom row, shape infered (right) by a single view image (left) using the encoding network. 
	Input images were segmented, binarized and resized to match the network input.}
\end{figure}

\subsubsection{Unsupervised shape and viewpoint prediction}
Our method is also able to handle unsupervised prediction of shapes in 2D images.
Once trained, the 3D shape generator is capable of creating shapes from
a set of encodings $z \in \mathbb{R}^{201}$.
One application is to predict the encoding of the underlying 3D object given a single view image of the object.
We do so by using the \prgan's generator to produce a large number of
encoding-image pairs, then use the data to train a neural network
(called encoding network). In other words, we create a training set
that consists of images synthesized by the \prgan and the encodings
that generated them. The encoding network is fully connected, with 2
hidden layers, each with 512 neurons.
The input of the network is an image and the output is an encoding.
The last dimension of $z$ describes the view, and the first 200
dimensions describe the code of the shape, which allows us to further
reconstruct the 3D shape as a $32^3$ voxel grid. With the encoding
network, we can present to it a single view image, and it outputs the
shape code along with the viewing angle. 
Experimental results are shown in in Figure~\ref{fig:sp}. This whole
process constitutes a completely unsupervised approach to creating a
model that infers a 3D shape from a single image.

\begin{figure*}
  \newcommand{\fh}{0.19\linewidth}
  \begin{center}
  \setlength{\tabcolsep}{3pt}
  \begin{tabular}{ccc}
    \textbf{Input} & \textbf{Generated images} & \textbf{Generated shapes} \\
    \includegraphics[height=\fh]{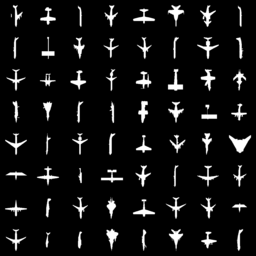} & 
    \includegraphics[height=\fh]{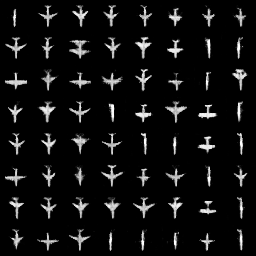} & 
    \includegraphics[height=\fh]{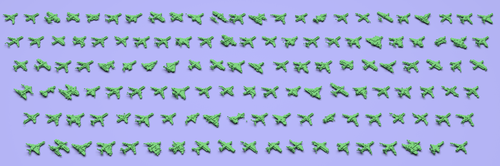} \\
    \includegraphics[height=\fh]{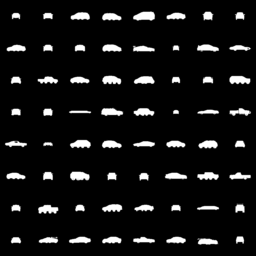} & 
    \includegraphics[height=\fh]{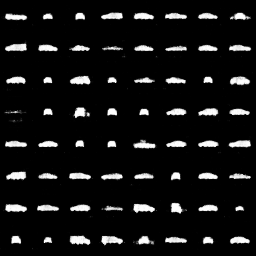} & 
    \includegraphics[height=\fh]{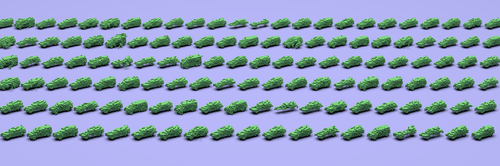} \\
    \includegraphics[height=\fh]{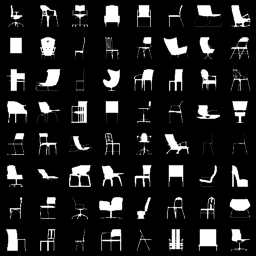} & 
    \includegraphics[height=\fh]{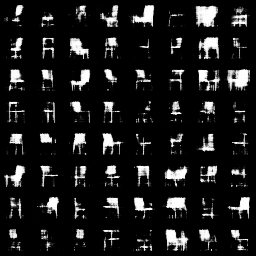} & 
    \includegraphics[height=\fh]{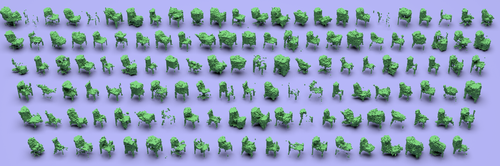} \\
    \includegraphics[height=\fh]{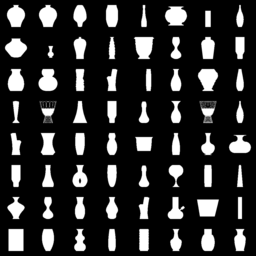} & 
    \includegraphics[height=\fh]{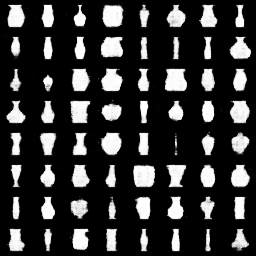} & 
    \includegraphics[height=\fh]{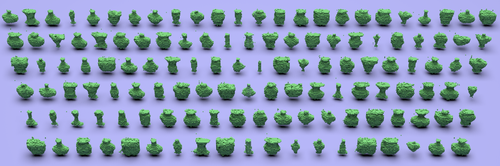} \\
    \includegraphics[height=\fh]{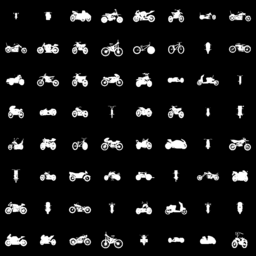} & 
    \includegraphics[height=\fh]{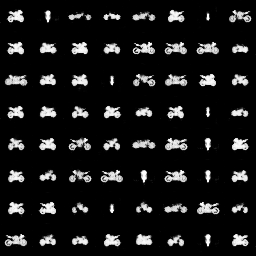} & 
    \includegraphics[height=\fh]{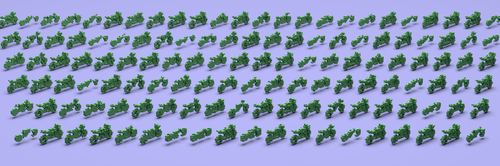} \\
    \includegraphics[height=\fh]{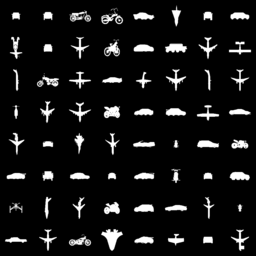} & 
    \includegraphics[height=\fh]{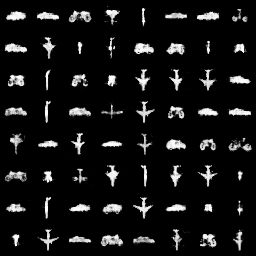} & 
    \includegraphics[height=\fh]{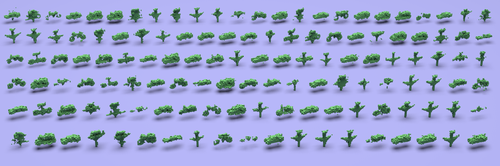} \\
  \end{tabular}
  \end{center}
  \vspace{-6pt}
  \caption{\label{fig:generate-shapes} Results for 3D shape induction using \prgans. From top to bottom we show results for airplane, car, chair, vase, motorbike, and a 'mixed' category obtained by combining training images from airplane, car, and motorbike. At each row, we show on the left 64 randomly sampled images from the input data to the algorithm, on the right 128 sampled 3D shapes from \prgan, and in the middle 64 sampled images after the projection module is applied to the generated 3D shapes. The model is able to induce a rich 3D shape distribution for each category. The mixed-category produces reasonable 3D shapes across all three combined categories. Zoom in to see details.}
\end{figure*}

\begin{figure*}[t]
\centering
\setlength{\tabcolsep}{0pt}
\begin{tabular}{cccccccccc}
\includegraphics[width=.1\linewidth]{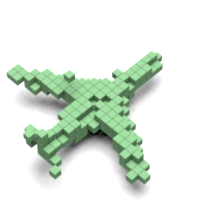} &
\includegraphics[width=.1\linewidth]{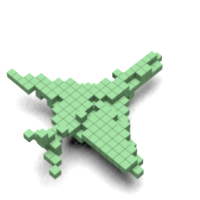} &
\includegraphics[width=.1\linewidth]{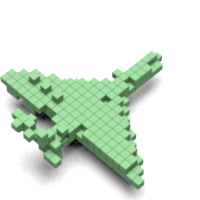} &
\includegraphics[width=.1\linewidth]{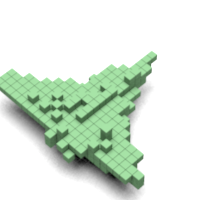} &
\includegraphics[width=.1\linewidth]{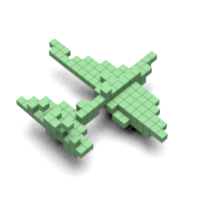} &
\includegraphics[width=.1\linewidth]{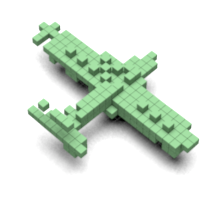} &
\includegraphics[width=.1\linewidth]{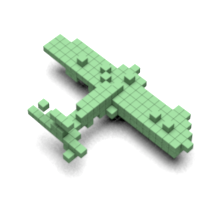} &
\includegraphics[width=.1\linewidth]{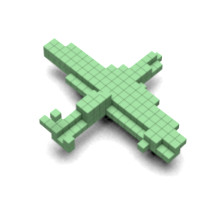} &
\includegraphics[width=.1\linewidth]{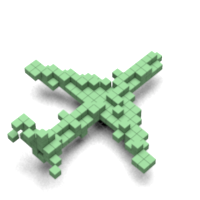} &
\includegraphics[width=.1\linewidth]{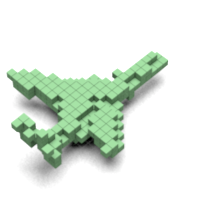} \\
\includegraphics[width=.1\linewidth]{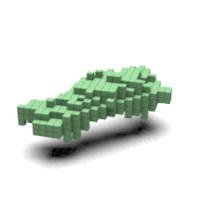} &
\includegraphics[width=.1\linewidth]{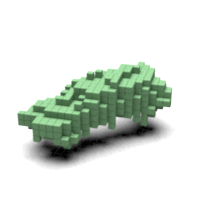} &
\includegraphics[width=.1\linewidth]{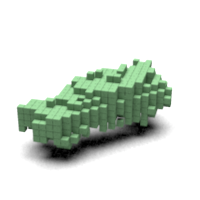} &
\includegraphics[width=.1\linewidth]{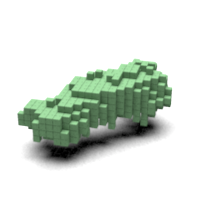} &
\includegraphics[width=.1\linewidth]{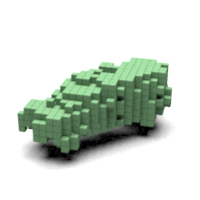} &
\includegraphics[width=.1\linewidth]{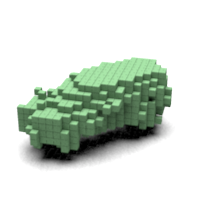} &
\includegraphics[width=.1\linewidth]{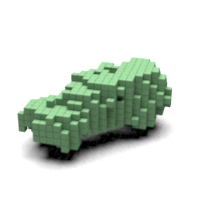} &
\includegraphics[width=.1\linewidth]{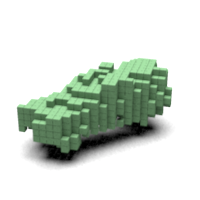} &
\includegraphics[width=.1\linewidth]{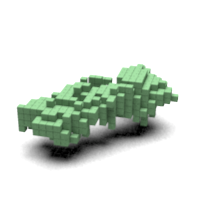} &
\includegraphics[width=.1\linewidth]{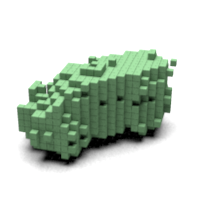} \\
\includegraphics[width=.1\linewidth]{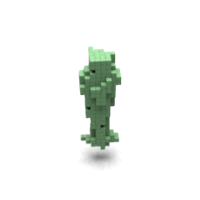} &
\includegraphics[width=.1\linewidth]{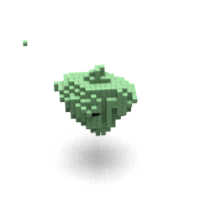} &
\includegraphics[width=.1\linewidth]{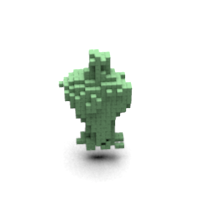} &
\includegraphics[width=.1\linewidth]{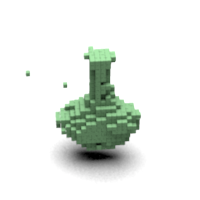} &
\includegraphics[width=.1\linewidth]{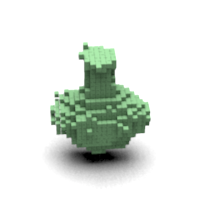} &
\includegraphics[width=.1\linewidth]{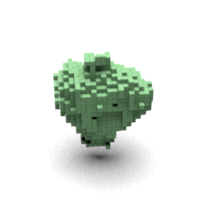} &
\includegraphics[width=.1\linewidth]{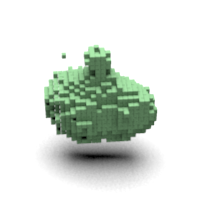} &
\includegraphics[width=.1\linewidth]{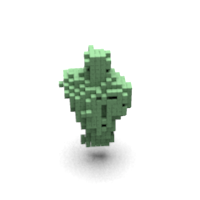} &
\includegraphics[width=.1\linewidth]{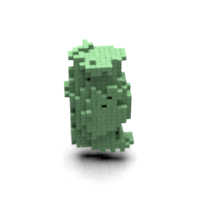} &
\includegraphics[width=.1\linewidth]{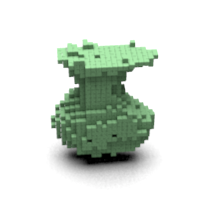} \\
\includegraphics[width=.1\linewidth]{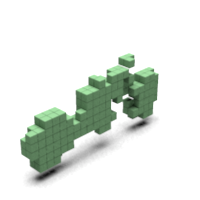} &
\includegraphics[width=.1\linewidth]{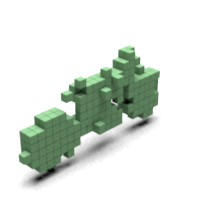} &
\includegraphics[width=.1\linewidth]{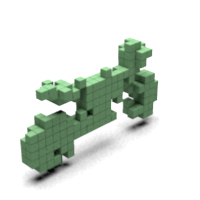} &
\includegraphics[width=.1\linewidth]{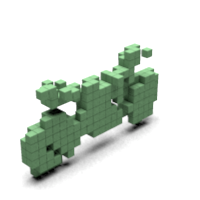} &
\includegraphics[width=.1\linewidth]{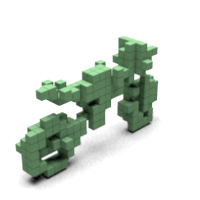} &
\includegraphics[width=.1\linewidth]{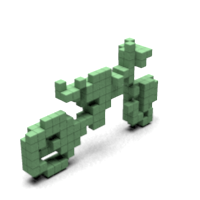} &
\includegraphics[width=.1\linewidth]{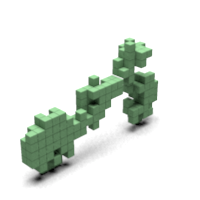} &
\includegraphics[width=.1\linewidth]{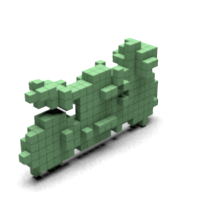} &
\includegraphics[width=.1\linewidth]{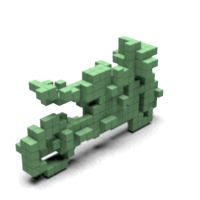} &
\includegraphics[width=.1\linewidth]{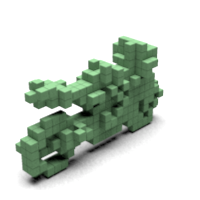} 
\end{tabular}
\caption{\label{fig:example-samples} A variety of 3D shapes generated by  \prgan trained on 2D views of (from the top row to the bottom row) airplanes, cars, vases, and bikes. These examples are chosen from the gallery in Figure~\ref{fig:generate-shapes} and demonstrate the quality and diversity of the generated shapes.}\end{figure*}

\begin{figure}[t]
\centering
\includegraphics[width=0.9\linewidth]{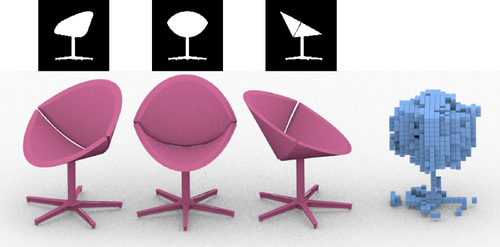}
\caption{\label{fig:failure} Our method is unable to capture the concave interior structures in this chair shape. The pink shapes show the original shape used to generate the projected training data, shown by the three binary images on the top (in high resolution). The blue voxel representation is the inferred shape by our model. Notice the lack of internal structure.}
\end{figure}

\subsubsection{Visualizations across categories} 
Our method is able to generate 3D shapes for a wide range of
categories. 
Figure~\ref{fig:generate-shapes} show a gallery of results,
including airplanes, car, chairs, vases, motorbikes. For each category
we show 64 randomly sampled training images, 64 generated images from
\prgan, and renderings of 128 generated 3D shapes (produced by randomly
sampling the 200-d input vector of the generator). 
One remarkable property is that the generator produces 3D shapes in a
consistent horizontal and vertical axes, even though the training data
is only consistently oriented along the vertical axis. Our hypothesis
for this is that the generator finds it more efficient to generate
shapes in a consistent manner by sharing parts across
models. Figure~\ref{fig:example-samples} shows selected examples from
Figure~\ref{fig:generate-shapes} that demonstrates the quality and
diversity of the generated shapes. 

The last row in Figure~\ref{fig:generate-shapes} shows an example of a
``mixed'' category, where the training images combine the three
categories of airplane, car, and motorbike. The same \prgan network is
used to learn the shape distributions. Results show that \prgan learns
to represent all three categories well, without any additional
supervision.

\subsection{Failure cases} 
Compared to 3D-GANs, the proposed \prgan models cannot discover structures that are hidden
due to occlusions from all views. For example, it fails to discover that some chairs have concave interiors and the generator simply fills these since it does not change the silhouette from any view as we can see at Figure
\ref{fig:failure}. 
However, this is a natural drawback of view-based approaches since
some 3D ambiguities cannot be resolved (\eg, necker cubes) without
relying on other cues. Despite this, one advantage over 3D-GAN is that
our model does not require consistently aligned 3D shapes since it
reasons over viewpoints.

\section{Improving \prgan~with richer supervision}\label{s:discussion}
In this section we show how the generative models can be improved to
support higher resolution 3D shapes and by incoporating richer forms
of view-based supervision.

\subsection{Higher-resolution models.} 
We extend the vanilla \prgan model to handle higher resolution volumes.
There two key modifications.
First, we replace the transposed convolutions in the generator by 
trilinear upsampling followed by a 3D convolutional layer.
In our experiments, we noticed that this modification led to smoother shapes
with less artifacts.
This fact was also verified for image generators~\cite{odena2016deconvolution}.
Second, we add a feature matching component to the generator objective.
This component acts by minimizing the difference between features computed by
the discriminator from real and fake images.
More precisely, the feature matching loss can be defined as:

\begin{equation}
	\mathcal{L}_{FM}(G, D) = \norm{\mathbb{E}_{x \sim \mathcal{D}}[D_k(x)] - 
								 \mathbb{E}_{z \sim \mathcal{N}(0, I)}[D_k(G(z))]}_2^2
\end{equation}
where $D_k(x)$ are the features from the $k$th layer of the discriminator when given
an input $x$.
In our experiments we define $k$ to be the last convolutional layer of the discriminator.
We empirically verified that this component promotes diversity in the 
generated samples and makes the training more stable.

\subsection{Using multiple cues for shape reasoning} 
Our approach currently only relies on binary silhouettes for
estimating the shape. 
This contributes to the lack of geometric
details, which can be improved by incorporating visual cues. 
One
strategy is to train a more powerful differentiable function
approximation, \eg, a convolutional network, to replicate the
sophisticated rendering pipelines developed by the computer graphics
community. 
Once trained, the resulting \emph{neural renderer} could be a plug-in
replacement for the \emph{projection module} in the \prgan framework.
This would allow the ability to use collections of
realistically-shaded images for inferring probabilistic models of 3D
shapes and other properties.
Recent work on screen-space shading using convnets are
promising~\cite{nalbach2016deep}.

Alternatively, we explore designing differentiable projection operators
that do not rely on training procedures.
This choice fits well int \prgan formulation as it does not
rely on 3D supervision for training any part of the model.
In this section, we present differentiable operators to render
depth images and semantic segmentation maps.
We demonstrate that the extra supervision enables generating more
accurate 3D shapes and allows relaxing the prior assumption
on viewpoint distribution.

\begin{figure*}[t]
\centering
\includegraphics[width=\linewidth]{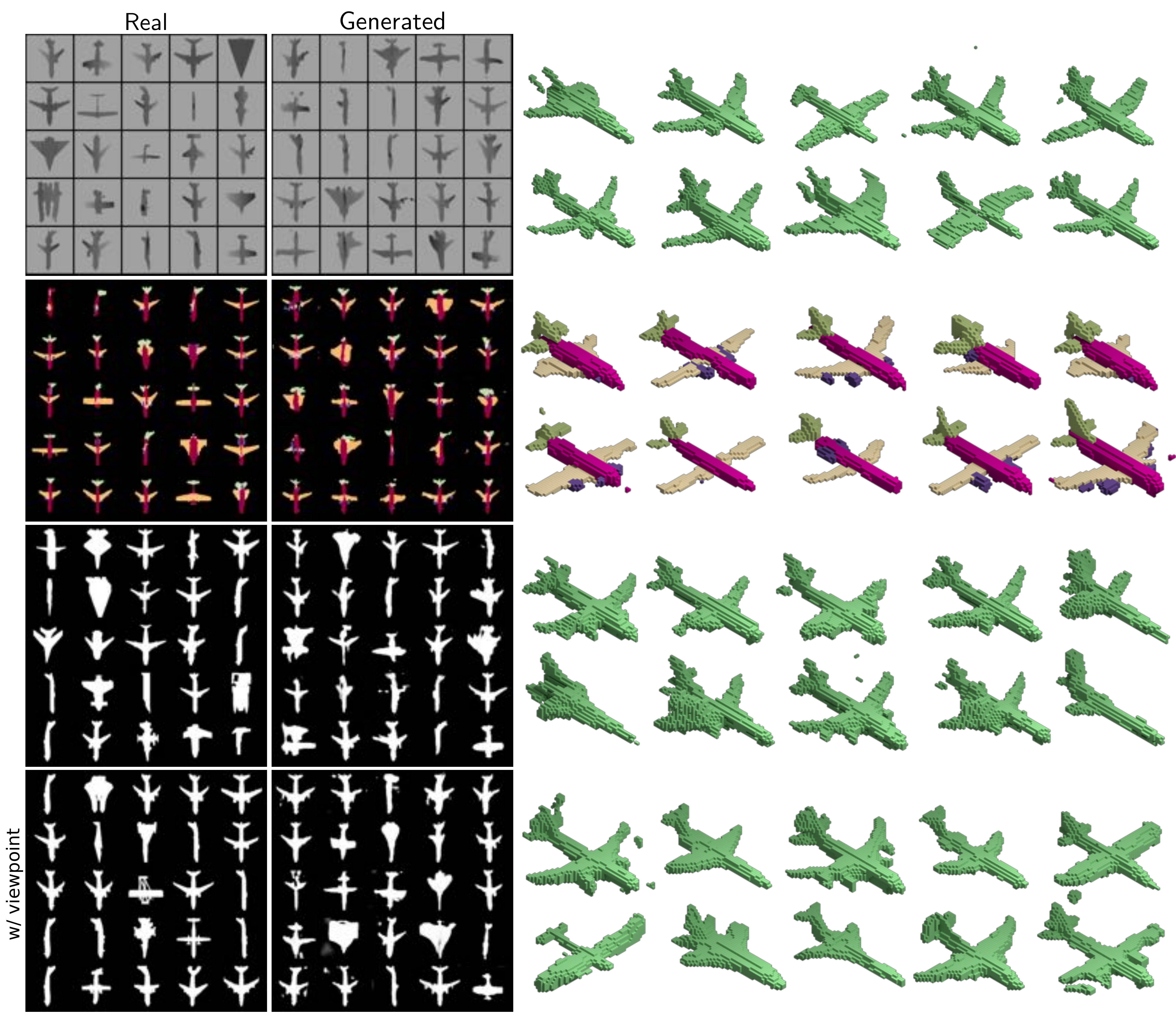}
\caption{\label{fig:newprojs} Shapes generated using new part segmentations and depth maps.
	From top to bottom, results using depth images, images with part segmentation, silhouettes and silhouettes annotated with viewpoints.
	Models trained with images containing additional visual cues are able to generated more accurate shapes. 
	Similarly, viewpoint annotation also helps.
	Notice that shapes generated from images with part annotation are able to generate part-annotated 3D shapes, highlighted by different colors.}
\end{figure*}

\paragraph{Learning from depth images.}
Our framework can be adapted to learn from depth images instead of binary images.
This is done by replacing the binary projection operator $Pr$ to one that can be used to generate
depth images.
We follow an approach inspired by the binary projection.
First, we define an accessibility function $A(V, \phi, c)$ that describes whether a given voxel $c$ inside the grid $V$ is visible, 
when seen from a view $\phi$:
\begin{equation}
A(V, \phi, i, j, k) = \exp\bigg\{-\tau \sum_{l=1}^{k-1} V_\phi(i,j,l) \bigg\}.
\end{equation}

Intuitively, we are incrementally accumulating the occupancy (from the first voxel on the line of sight) as we traverse the voxel grid 
instead of summing all voxels on entire the line of sight. 
If voxels on the path from the first to the current voxel are all
empty, the value of $A$ is 1 (indicating the current voxel is
``accessible'' to the view $\phi$). 
If there is at least one non-empty voxel on the path, the value of A
will be close to 0 (indicating this voxel is inaccessible).

Using $A$, we can define the depth value of a pixel in the projected image
as the line integral of A along the line of sight: $Pr^{D}_\phi(i,j, V)=\sum_{k}A(V,\phi,i,j,k)$. 
This operation computes the number of accessible voxels from a particular direction $\phi$, which
corresponds to the distance of the surface seen in $(i,j)$ to the camera.
Finally, we apply a smooth map to the previous operation in order to have depth values in the range $[0, 1]$.
Thus, the projection module is defined as:
\begin{equation}
\label{eq:expdepth}
	Pr^{D}_\phi((i,j),V) = 1 - \exp\bigg\{-\sum_{k}A(V,\phi,i,j,k)\bigg\}.
\end{equation}

\paragraph{Learning from part segmentations.}

We also explore learning 3D shapes from sets of images with dense semantic annotation.
Similarly to the depth projection, we modify our projection operator to enable generation
of images whose pixels correspond to the label of particular class (or none if there is no object).
In this case, the output of the generator is multi-channel voxel grid 
$V:\mathbb{Z}^3 \times C \rightarrow [0,1] \in \mathbb{R}$, where $C$ is the number of parts present
in a particular object category.

Let $G$ to be the aggregated occupancy grid defined as $G=\sum_{c=1}^C V(i,j,k,c)$.
The semantic projection operator $Pr^{S}_\phi((i,j,c), V)$ is defined as:
\vspace{-10pt}
\begin{equation}
  Pr^{S}_\phi\left((i, j, c), V\right) = 1 - \exp\bigg\{\sum_k V_\phi(i,j,k,c) A(G_\phi, i, j, k)\bigg\},
\end{equation}
where $A$ is the accessibility operator defined previously.
Intuitively, $A(G, \phi)$ encodes if a particular voxel is visible from a viewpoint $\phi$.
When we multiply the visibility computed with the aggregated occupancy grid by the value of
a specific channel $c$ in $V$, we generate a volume that contains visibility information per part.
Finally, we take the line integral along the line of sight to generate the final image.
Examples of images and shapes generated by this operator can be seen in Figure~\ref{fig:newprojs}.

\paragraph{Learning with viewpoint annotation.}
\label{s:viewpoint}

We also experiment with the less challenging setup where our model has access to viewpoint information
of every training image.
Notice that this problem is different from \cite{nmr,yan2016perspective}, since we still do not know
which images correspond to the same object.
Thus, multi-view losses are not a viable alternative.
Our model is able to leverage viewpoint annotation by using conditional discriminators.
The conditional discriminator has the same architecture as the vanilla discriminator but the input image is modified to contain
its corresponding viewpoint annotation.
This annotation is represented by an one-hot encoding concatenated to every pixel in the image.
For example, if a binary image from a dataset with shapes rendered from 8 viewpoints will be represented as a 9-channel image.
This procedure is done for images generated by our generated and images coming from the dataset.

\begin{table}
\centering
\setlength{\tabcolsep}{4pt}
\begin{tabular}{c|c|c|c|c}
Model  & Supervision & $\mathcal{D} \rightarrow G(z)$ & $G(z) \rightarrow \mathcal{D}$
  & Avg.  \\ 
\hline
\prgan & Silhouette  & 0.442   & 0.400   & 0.421 \\ 
\prgan & Silhouette + View & 0.439  & 0.431  & 0.435 \\
\prgan & Depth & 0.497  & 0.448    & 0.472 \\
\prgan & Part Segmentation & 0.496  & 0.507  & 0.502 \\ 
\hline
3D-GAN & Volumetric & 0.538 &0.530  &0.534\\
\end{tabular}
\caption{\label{tab:newcomp} Quantitative comparison between models
  trained with different projection operators. The Chamfer similarity
  under the volumetic intersection over union (IoU) is shown for
  \prgan trained with varying amounts of supervision and a 3D-GAN
  trained with volumetric supervision. 
  The metric (higher the better) indicates that
  \prgan with richer supervision are better and approaches the
  quality of 3D-GAN.}
\end{table}

\subsection{Experiments}

\paragraph{Setup.}
We generate training images using airplanes from the ShapeNet part segmentation dataset~\cite{shapenet}.
Those shapes have their surface densely annotated as belonging to one of four parts:
body, wing, tail or engine.
We render those shapes using the same viewpoint configuration described in Section~\ref{s:experiments}.
However, in this scenario we use $64\times 64$ images instead of $32\times 32$.
The models are rendered as binary silhouettes, depth maps and part segmentation masks.
We train a high resolution \prgan model for every set of rendered images using the corresponding
projection operator.
Each model is trained for 50 epochs and trained with Adam optimizer.
We use a learning rate of $2.5 \times 10^{-3}$ for the generator and
$2 \times 10^{-5}$ for the discriminator.

\paragraph{Evaluation.}
The models trained with different visual clues are evaluated through the following metric:
\begin{equation}
	\frac{1}{|\mathcal{D}|}\sum_{x \in \mathcal{D}} \min_{g \in \mathcal{G}} IoU(x, g) +
	\frac{1}{|\mathcal{G}|}\sum_{g \in \mathcal{G}} \min_{x \in \mathcal{D}} IoU(x, g) 
	\label{eq:metric}
\end{equation}
where $IoU$ corresponds to intersection over union, $\mathcal{G}$ is a set of generated shapes and
$\mathcal{D}$ is a set of shapes from the training data.
In our setup, both $\mathcal{G}$ and $\mathcal{D}$ contain 512 shapes.
Shapes in $\mathcal{D}$ are randomly sampled from the same dataset that originated the images,
whereas shapes in $\mathcal{G}$ are generated through $G(z)$.
Noticeably, the shapes generated by \prgan do not have the same orientation as the shapes
in $\mathcal{D}$ but are consistently oriented among themselves.
Thus, before computing Equation~\ref{eq:metric}, we select one of 8 possible transformations that minimizes
$IoU$ -- there are 8 rendering viewpoints in the training set.
Additionally, the components in Equation~\ref{eq:metric} indicate two different aspects:
the first term ($\mathcal{D}\rightarrow G(z)$) indicates how the variety in the dataset is covered whereas 
the second term ($G(z)\rightarrow \mathcal{D}$) indicates how accurate the generated shapes are.
A comparison between models trained with different projection operators can be seen in
Table~\ref{tab:newcomp}.
The model trained with part segmentation clues yields the best results.
As expected, using only silhouettes leads to worse results in both metrics and adding viewpoint
supervision improves upon this baseline.
Interestingly, depth and part segmentation supervision clues lead to models that generate shapes
with similar variety (similar $\mathcal{D}\rightarrow G(z)$).
However, shapes generated from models using part segmentation clues are more similar to the ones
in the dataset (higher $G(z)\rightarrow \mathcal{D}$).

\subsection{Learning from real images} 
Our approach can be extended to learning 3D shapes from real images by
applying an existing approach for segmentation such
as~\cite{long2015fully}. However, the assumption that the viewpoints
are uniformly distributed over the viewing sphere may not hold. In
this situation, one can either learn a distribution over viewpoints by
mapping a few dimensions of the input code $z$ to a distribution over
viewpoints $(\theta,\phi)$ using a multi-layer network. More
generally, one can also learn a distribution over a full set of camera
parameters. An alternative is learn a conditional model where the
viewpoint is provided as input to the algorithm, 
like the model we experimented in Section~\ref{s:viewpoint}. 
This extra annotation may be obtained using a generic viewpoint estimator such
as~\cite{tulsiani2015pose,su2015render}.

\section{Conclusion}\label{s:conclusion}
We proposed a framework for infering 3D shape distributions from 2D
shape collections by agumenting a convnet-based 3D shape generator
with a projection module. This compliments exisiting approches for
non-rigid SfM since these models can be trained without prior
knowledge about the shape family, and can generalize to categories
with variable structure. We showed that our models can infer 3D shapes
for a wide range of categories, and can be used to infer shape and
viewpoint from a single image in a completely unsupervised manner. We
believe that the idea of using a differentiable render to infer
distributions over unobserved scene properties from images can be
applied to other problems.

\paragraph{Acknowledgement:} This research was supported in part by the
NSF grants 1617917, 1749833, 1423082 and 1661259. 
The experiments were performed using equipment obtained under a grant
from the Collaborative R\&D Fund managed by the Massachusetts Tech
Collaborative.

\bibliographystyle{ieee_fullname}
\bibliography{prgan-ijcv}

\end{document}